\documentclass[letterpaper, 10 pt, conference]{ieeeconf}  % Comment this line out if you need a4paper

\IEEEoverridecommandlockouts                              % This command is only needed if 
\usepackage{graphicx}
\usepackage{color}
\usepackage{amsfonts}
\usepackage{amsmath}
\usepackage{algorithm}
\usepackage{algpseudocode}
\usepackage{url}
\usepackage{tabularx}
\usepackage{multirow} 
\usepackage{booktabs} 

\title{\LARGE \bf
Autonomous Apple Fruitlet Sizing with Next Best View Planning
}

\author{Harry Freeman and George Kantor% <-this % stops a space
\thanks{H. Freeman and G. Kantor are with Carnegie Mellon University, Pittsburgh PA, USA \texttt{\{hfreeman, kantor\}@cs.cmu.edu}}
}

\begin{document}
\maketitle
\thispagestyle{empty}
\pagestyle{empty}

%%%%%%%%%%%%%%%%%%%%%%%%%%%%%%%%%%%%%%%%%%%%%%%%%%%%%%%%%%%%%%%%%%%%%%%%%%%%%%%%
\begin{abstract}
In this paper, we present a next-best-view planning approach to autonomously size apple fruitlets. State-of-the-art viewpoint planners in agriculture are designed to size large and more sparsely populated fruit. They rely on lower resolution maps and sizing methods that do not generalize to smaller fruit sizes. To overcome these limitations, our method combines viewpoint sampling around semantically labeled regions of interest, along with an attention-guided information gain mechanism to more strategically select viewpoints that target the small fruits' volume. Additionally, we integrate a dual-map representation of the environment that is able to both speed up expensive ray casting operations and maintain the high occupancy resolution required to informatively plan around the fruit. When sizing, a robust estimation and graph clustering approach is introduced to associate fruit detections across images. Through simulated experiments, we demonstrate that our viewpoint planner improves sizing accuracy compared to state of the art and ablations. We also provide quantitative results on data collected by a real robotic system in the field.
\end{abstract}

%%%%%%%%%%%%%%%%%%%%%%%%%%%%%%%%%%%%%%%%%%%%%%%%%%%%%%%%%%%%%%%%%%%%%%%%%%%%%%%%

%%%%%%%%%%%%%%%%%%%%%%%%%%%%%%%%%%%%%%%%%%%%%%%%%%%%%%%%%%%%%%%%%%%%%%%%%%%%%%%%
\section{Introduction}

With population growth, climate change, and labor shortages on the rise, the global food supply is continuously facing increasing pressure. Agriculturalists are actively looking to integrate robotics-based solutions to reduce the time and effort required to perform labor-intensive tasks. Among these solutions is autonomous crop inspection, which is integral for precision agriculture. This is because it enables farmers to make smarter and more informed decisions for a variety of downstream processes, including quality control, disease and pest management, pruning, and harvesting. Apple growers, for example, size apple fruitlets\footnote{A fruitlet is a young apple formed on the tree shortly after pollination} in order to determine when to apply chemical thinners to their crops in order to maintain a more consistent yearly harvest~\cite{greene2013development}. The current practice of manually obtaining sizes is labor-intensive, as calipers are used to record the diameters of approximately 500 fruitlets per varietal at a minimum of twice per thinning application. As a result, apple growers are looking to adopt autonomous solutions to improve efficiency and reduce cost.

\begin{figure}[!t]
    \centering
    \includegraphics[width= 1\linewidth]{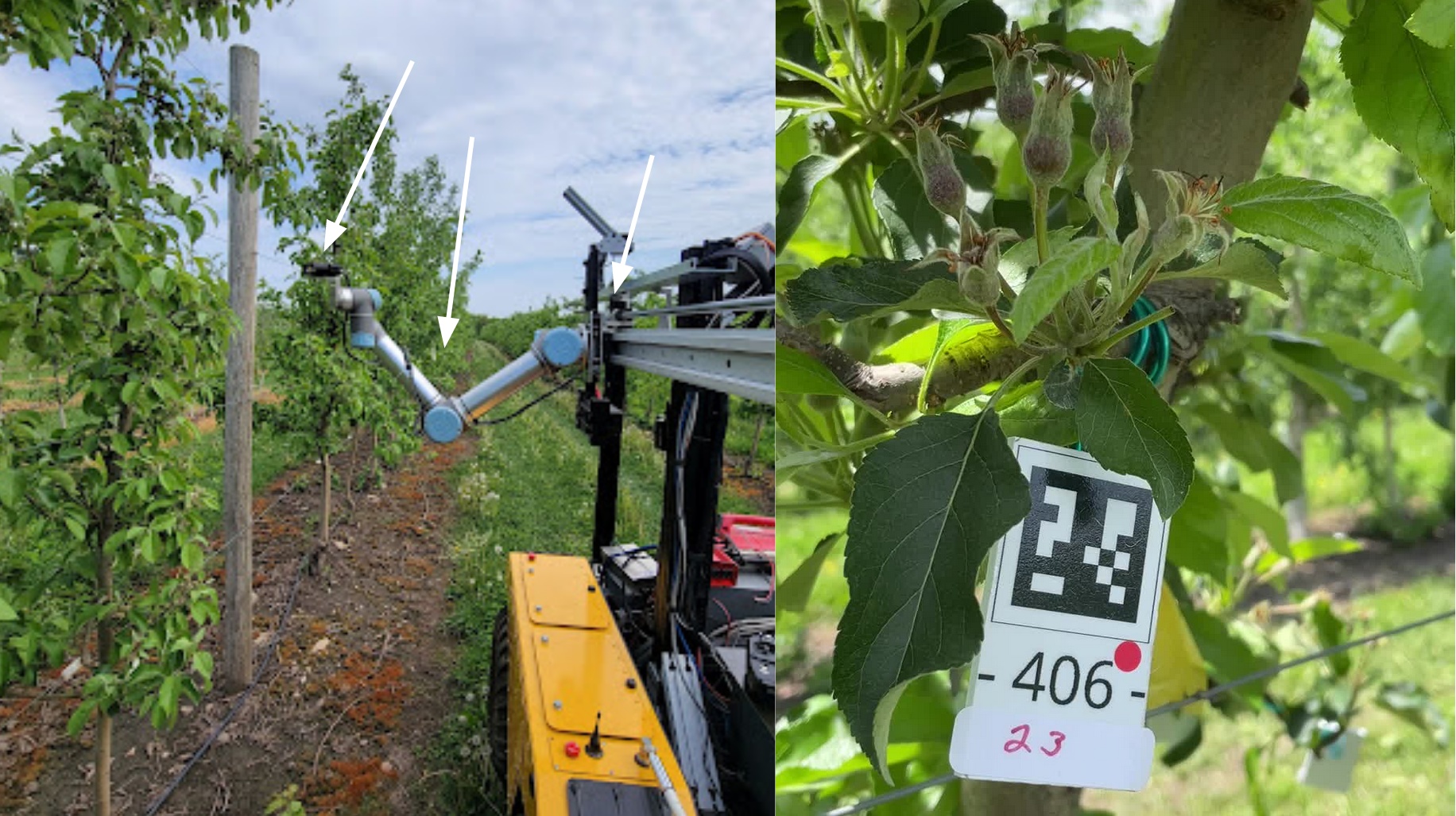}
    \put(-197, 130){\scriptsize Stereo Camera}
    \put(-174, 118){\scriptsize UR5}
    \put(-156, 114){\scriptsize Linear Slider}
    \vspace{-6pt}
    \caption{Left: Stereo camera attached to a 7 DoF robotic arm. Right: Example of a fruitlet cluster with AprilTag hung in close proximity for identification.}
    \label{fig:bb_and_fruitlet}
    \vspace{-20pt}
\end{figure}

Sizing fruit requires accurate spatial reasoning about the plants. This is often difficult to obtain, as many fruits grow in occluded environments. In the case of apple fruitlets (Fig.~\ref{fig:bb_and_fruitlet}), the fruits are only 7mm-14mm in diameter, they are surrounded by branches and leaves, and often fruitlets in the same cluster\footnote{A cluster is a group of fruitlets that grow out of the same bud} will occlude one another. Therefore, it is insufficient to use fixed camera systems attached to either stationary~\cite{fixed_apple} or mobile~\cite{taim} bases, as images of the fruit cannot be consistently captured. It is much more reliable to use active perception-based methods, where mobile cameras attached to robotic arms are free to move around, avoid occlusions, and capture more semantically meaningful images. This is commonly achieved using next-best-view (NBV) planning, where information stored inside a volumetric map is used to determine the next best camera pose that maximizes information gain.

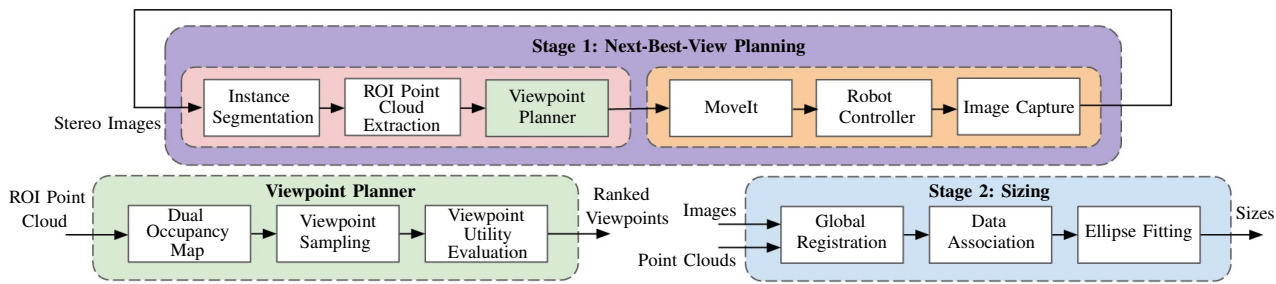
\begin{figure*}[t]
    \centering
    \includegraphics[width=0.9\linewidth]{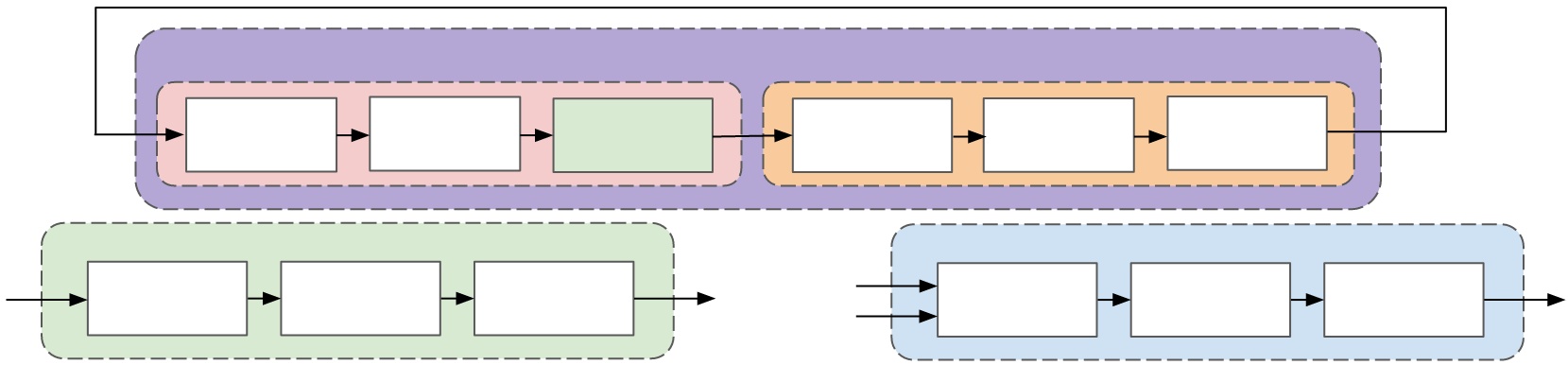}
    \put(-277, 88){\scriptsize \textbf{Stage 1: Next-Best-View Planning}}
    \put(-378, 33){\scriptsize \textbf{Viewpoint Planner}}
    \put(-127, 33){\scriptsize \textbf{Stage 2: Sizing}}
    \put(-458, 59){\scriptsize Stereo Images}
    \put(-392, 69){\scriptsize Instance}
    \put(-398, 61){\scriptsize Segmentation}
    \put(-341, 71){\scriptsize ROI Point}
    \put(-335, 65){\scriptsize Cloud}
    \put(-341, 58){\scriptsize Extraction}
    \put(-286, 69){\scriptsize Viewpoint}
    \put(-282, 61){\scriptsize Planner}
    \put(-212, 65){\scriptsize MoveIt}
    \put(-158, 69){\scriptsize Robot}
    \put(-161, 61){\scriptsize Controller}
    \put(-114, 65){\scriptsize Image Capture}
    \put(-475, 31){\scriptsize ROI Point}
    \put(-470, 22){\scriptsize Cloud}
    \put(-415, 24){\scriptsize Dual}
    \put(-423, 18){\scriptsize Occupancy}
    \put(-414, 11){\scriptsize Map}
    \put(-365, 22){\scriptsize Viewpoint}
    \put(-365, 14){\scriptsize Sampling}
    \put(-309, 24){\scriptsize Viewpoint}
    \put(-303, 17){\scriptsize Utility}
    \put(-309, 10){\scriptsize Evaluation}
    \put(-254, 32){\scriptsize Ranked}
    \put(-257, 23){\scriptsize Viewpoints}
    \put(-220, 26){\scriptsize Images}
    \put(-237, 7){\scriptsize Point Clouds}
    \put(-170, 22){\scriptsize Global}
    \put(-177, 14){\scriptsize Registration}
    \put(-111, 22){\scriptsize Data}
    \put(-120, 14){\scriptsize Association}
    \put(-68, 18){\scriptsize Ellipse Fitting}
    \put(-11, 25){\scriptsize Sizes}
    \vspace{-8pt}
    \caption{Overview of our next-best-view planning and sizing pipeline.
    \label{fig:nbv_full_pipeline}}
    \vspace{-15pt}
\end{figure*}

However, state-of-the-art NBV planners in agriculture are dedicated towards sizing larger fruit that are sparsely populated. They rely on expensive ray casting operations and shape completion methods that do not generalize to smaller fruit sizes. To overcome these limitations, we present an NBV planning approach that utilizes regions of interest for viewpoint sampling and an attention-guided mechanism for calculating information gain. In addition, we integrate a dual-map representation of the environment that significantly speeds up ray casting while maintaining finer occupancy information. To address the challenges of data association in the presence of wind and sensor error, we introduce a robust estimation and graph clustering approach. We demonstrate that our planning method improves sizing accuracy compared to a state-of-the-art viewpoint planner used in agriculture. Although our target application is sizing apple fruitlets, our method could be adapted to size other fruits and plants of comparable size. Our main contributions are

\begin{itemize}
    \item A novel next-best-view planner that uses coarse and fine occupancy maps, region of interest targeted viewpoint sampling, and an attention-guided information gain metric to capture images of smaller fruit.
    \item A robust estimation and graph clustering approach to associate fruit detections across images in the presence of wind and sensor error.
    \item Quantitative evaluation on data collected by a real robotic system in a commercial apple orchard.
\end{itemize}

% Comparable size

%%%%%%%%%%%%%%%%%%%%%%%%%%%%%%%%%%%%%%%%%%%%%%%%%%%%%%%%%%%%%%%%%%%%%%%%%%%%%%%%
\section{Related Work}

There have been previous works dedicated towards viewpoint planning in agriculture. Few approaches rely on sensor information without maintaining a volumetric map of the environment. Lehnert \textit{et al.}~\cite{3dmovetosee} use a 3D camera array to size a target fruit of interest. They find the next best view by computing a local gradient and determining the direction in which the visibility of the target increases. While this method is effective at avoiding local occlusions, it does not extend globally when multiple fruit need to be sized. Roy \textit{et al.}~\cite{counting_apples} present a viewpoint planning approach to count the number of apples on a tree branch. They determine the next camera pose by maximizing entropy over a set of world hypotheses informed by previous images. However, they make simplified assumptions about the relative positions and occlusion levels of the apples, which does not extend to smaller fruit that are more easily displaced and obscured by leaves.

More commonly, volumetric-based planning approaches have been adopted in agriculture, where occupancy information stored inside a volumetric map is used to guide the viewpoint planning process. The information gain for viewpoint candidates is often calculated by casting rays through the volumetric map. Wu \textit{et al.}~\cite{deep_learn_nbv} use the volumetric map as input into a deep learning-based Point Completion Network~\cite{PCN} to estimate the unobserved points of a plant. The points are then used to sample viewpoint candidates. While deep learning-based approaches work for phenotyping plants, the absence of real and synthetic datasets makes them unadaptable to sizing uncommon fruit, such as apple fruitlets. Non-deep learning-based based methods often utilize spatial knowledge of semantically labeled regions of interest (ROIs). In the work of~\cite{dec_mcts}, detected ROIs in the form of segmented apples are used to estimate information gain. The information gain is used as input into a Decentralized Monte Carlo Tree Search~\cite{decmcts} planner that outputs a sequence of viewpoints for multiple robot arms. However, computing a sequence of viewpoints is computationally expensive. As well, viewpoints are locally sampled without utilizing the global map. Zaenker \textit{et al.} \cite{nbv_zaenker} address these issues by introducing ROI targeted sampling for NBV planning. Viewpoint targets are sampled from frontiers in the vicinity of known ROIs and are used to determine the next best view. They later integrate the work of~\cite{3dmovetosee} to combine local and global viewpoint planning to size capsicums~\cite{combglobloc}. While the computational cost is less, these methods still rely on expensive ray casting operations when calculating information gain, which are slow when performed at the resolution required to map millimeter-sized fruit. The authors of~\cite{dissim_zaenker} introduce a viewpoint dissimilarity metric based on autonomous shape completion to eliminate ray casting from the information gain formulation. However, shape completion methods tend to perform poorly on small fruit. This is because it is difficult to capture enough of the fruit's surface. As well, wind and sensor error have a more significant effect, making the tasks of point cloud registration between frames and fruit clustering much more challenging.

Burusa \textit{et al.}~\cite{burusa2022attentiondriven} propose an attention-guided NBV planning approach. Attention regions in the form of 3D bounding boxes are defined around different plant components and are used to restrict which voxels contribute to the information gain. The authors demonstrate that this attention mechanism leads to improved results with regards to both reconstruction accuracy and speed. However, the location and size of the attention regions were assumed to be prior knowledge, which is not the case when sizing fruit in unknown environments.
%As well, they did not consider ROI information when viewpoint sampling, and sample viewpoints on a cylindrical sector which was also assumed to be known beforehand.

%%%%%%%%%%%%%%%%%%%%%%%%%%%%%%%%%%%%%%%%%%%%%%%%%%%%%%%%%%%%%%%%%%%%%%%%%%%%%%%%
\section{System Overview}\label{sec:sys_overview}

An overview of our system can be seen in Fig.~\ref{fig:nbv_full_pipeline}. Our pipeline is composed of two stages. In the first stage, next-best-view planning is used to determine camera poses that improve spatial knowledge  of the scene. An AprilTag~\cite{apriltag} is hung in close proximity to each cluster for identification. Images are captured using an illumination-invariant flash stereo camera~\cite{flashcam2021} attached to the end of a 7 DoF robotic arm consisting of a UR5 and linear slider~\cite{silwal2021bumblebee}. Examples of our robotic system and tagged fruitlet cluster can be seen in Fig.~\ref{fig:bb_and_fruitlet}. When planning, the fruitlets are first segmented using a Mask-RCNN~\cite{maskrcnn2017} instance segmentation network, and disparities are extracted using RAFT-Stereo~\cite{lipson2021raft}. The segmentations and disparities are used to build a point cloud of the scene with ROI information. The viewpoint planner updates a dual-map representation of the environment, which is used to sample candidate viewpoints and determine the next best end-effector pose based on expected utility. A path is planned to the pose using the MoveIt framework~\cite{moveit}, which is executed by the robot controller. The process repeats until the specified planning duration is exceeded.

Sizing is performed once viewpoint planning is complete. To account for re-projection error resulting from sensor noise and wind, all point clouds are globally registered using a robust estimation method. Data association between images is then performed using Highly Connected Subgraph Clustering \cite{hcs} to account for outliers and spurious detections. Lastly, fruitlets are sized using a 2D photogrammetric method.

%%%%%%%%%%%%%%%%%%%%%%%%%%%%%%%%%%%%%%%%%%%%%%%%%%%%%%%%%%%%%%%%%%%%%%%%%%%%%%%%
\section{Viewpoint Planning}\label{sec:viewpoint_planning}

Our viewpoint planner combines ROI targeted sampling and attention mechanisms from previous works~\cite{nbv_zaenker, burusa2022attentiondriven}. We also utilize a dual-map representation of the environment in the form of coarse and fine octrees. This representation allows us to reduce the cost of ray casting while maintaining finer map resolutions within attention boundaries.

\subsection{Workspace and Sampling Tree Generation}\label{tree_gen}

Our viewpoint planner uses both workspace and sampling trees to generate viewpoint candidates. The workspace tree defines the valid end-effector positions the robot is able to reach, while the sampling tree identifies the regions of space that viewpoint targets should be sampled from. To generate the workspace tree, ten million randomly generated joint configurations were sampled in simulation.

\begin{figure}[ht]
    \centering
    \includegraphics[width=0.95\linewidth]{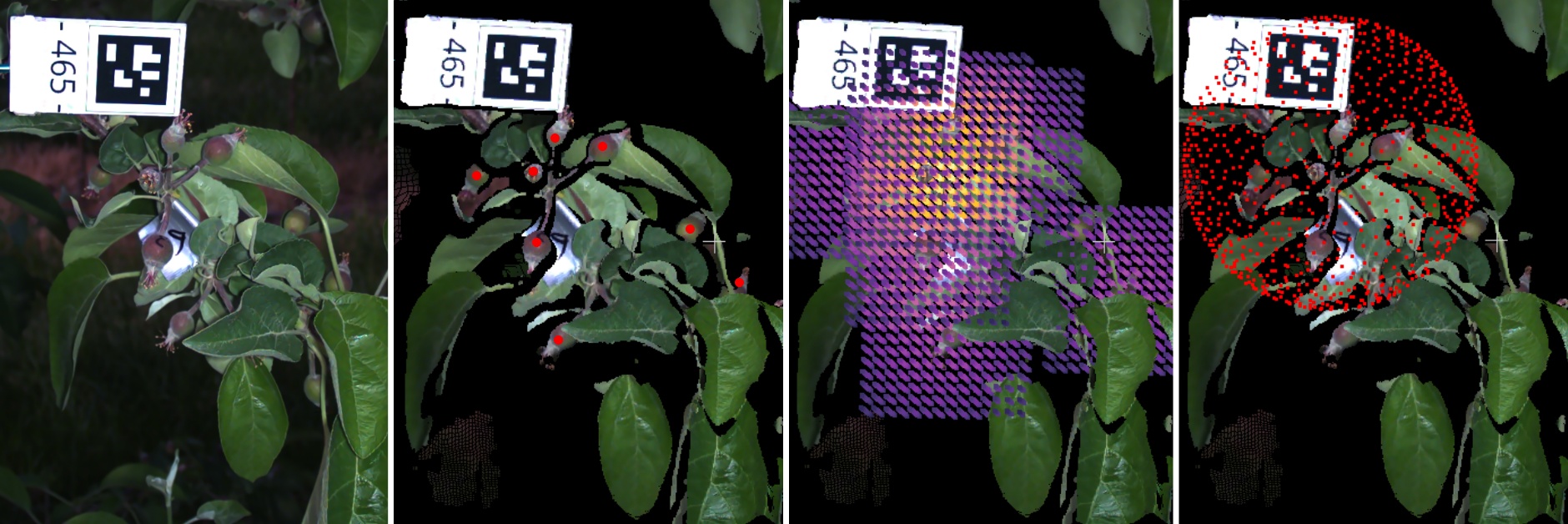}
    \put(-228, 5){\textcolor{white}{(a)}}
    \put(-168, 5){\textcolor{white}{(b)}}
    \put(-110, 5){\textcolor{white}{(c)}}
    \put(-53, 5){\textcolor{white}{(d)}}
    \vspace{-5pt}
    \caption{(a) Original RGB image with AprilTag near cluster. (b) Extracted point cloud with fruitlet centroids. (c) Density map created by smoothing volume around centroids. (d) Sampling tree created from local maxima.}
    \label{fig:nbv_cluster_extract}
    \vspace{-15pt}
\end{figure}
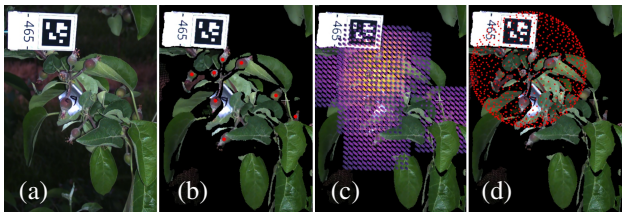

Because fruitlets are sized in clusters, we need a sampling tree large enough to encompass every fruitlet in the cluster, but not so large such that it includes additional fruitlets. To extract the sampling tree, we use a density-based clustering approach inspired from~\cite{sorghumscan}. First, an initial image is captured to detect the AprilTag and nearby fruitlets. We lift the fruitlet centers to 3D and create a density map by treating each centroid as a unit-impulse and applying a 3D Gaussian filter to smooth the space around it. Cluster centroids are found by extracting local maxima in the density map, and each fruitlet is assigned to the closest cluster based on Euclidean distance. The cluster whose local maximum is closest to the AprilTag is determined to be the target cluster, and the sampling tree is constructed by fitting a sphere around the corresponding fruitlets. An example of the process can be seen in Fig.~\ref{fig:nbv_cluster_extract}. We define the volume spanned by the sampling as the Attention Region. This is used to calculate information gain as described in Section~\ref{subsec:viewpoint_evaluation}.

\subsection{Dual-Map Representation}\label{subsec:dual_map}
 While libraries such as Octomap~\cite{octomap} have been designed to speed up expensive ray casting operations with optimized logic and multi-threading, these operations are still slow when performed at millimeter-level resolutions required to map small fruit. 
 % NBV planners in agriculture use expensive ray casting operations to update their volumetric maps and calculate information gain. While libraries such as Octomap~\cite{octomap} have been designed to speed up these operations with optimized logic and multi-threading, they are still slow when performed at millimeter-level resolutions required to map small fruit. %Using lower resolutions results in information loss due to the inability to accurately represent the fruit's occupancy.
To overcome this, we take advantage of the fact that ROIs are usually confined to certain regions of space. We do this by maintaining two maps of the environment: a coarse octree that stores occupancy information at lower resolution, and a fine octree that stores both occupancy and ROI information at higher resolution. The coarse octree spans the entire observation space and is used to approximate the occupancy of voxels and detect occlusions outside the Attention Region. In contrast, the fine octree is restricted to within the Attention Region, and is used to identify which voxels are in the vicinity of ROIs. This allows the planner to more efficiently evaluate occupancy and occlusions when viewpoint planning, while providing sufficient resolution to map the fruit. We use the OctoMap framework~\cite{octomap} and the ROI-extended implementation presented by Zaenker \textit{et al.}~\cite{nbv_zaenker} for the coarse and fine octrees respectively.

\begin{figure}[ht]
    \centering
    \includegraphics[width=0.8\linewidth]{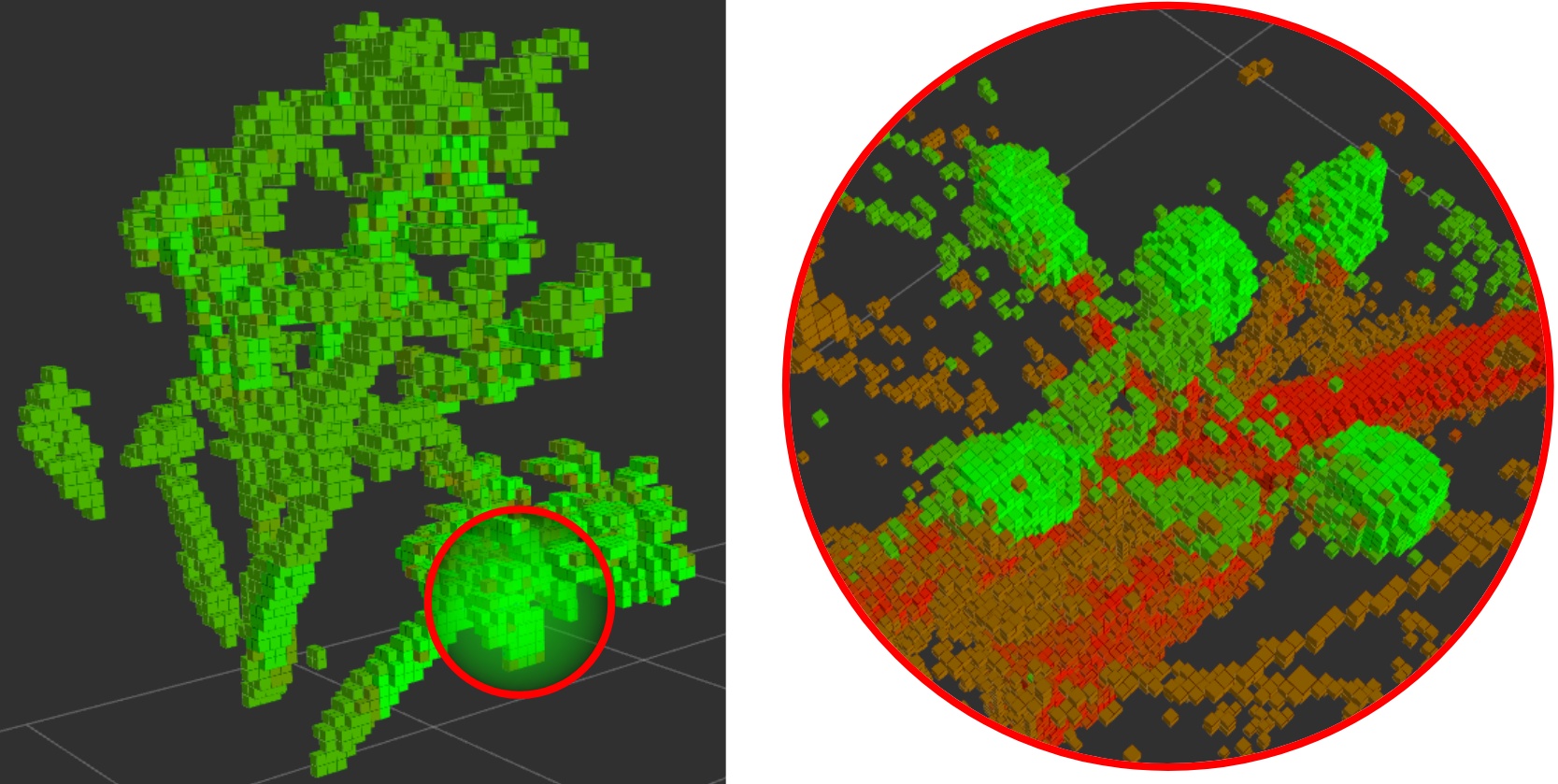}   
    \vspace{-5pt}
    \caption{Dual-map representation. Left: Coarse octree that stores occupancy information and spans the entire observation space. Right: Fine octree that stores occupancy and ROI information (green) within the Attention Region.}
    \label{fig:nbv_coarse_fine_vis}
    \vspace{-8pt}
\end{figure}

\begin{figure}[ht]
    \centering
    \includegraphics[width=0.95\linewidth]{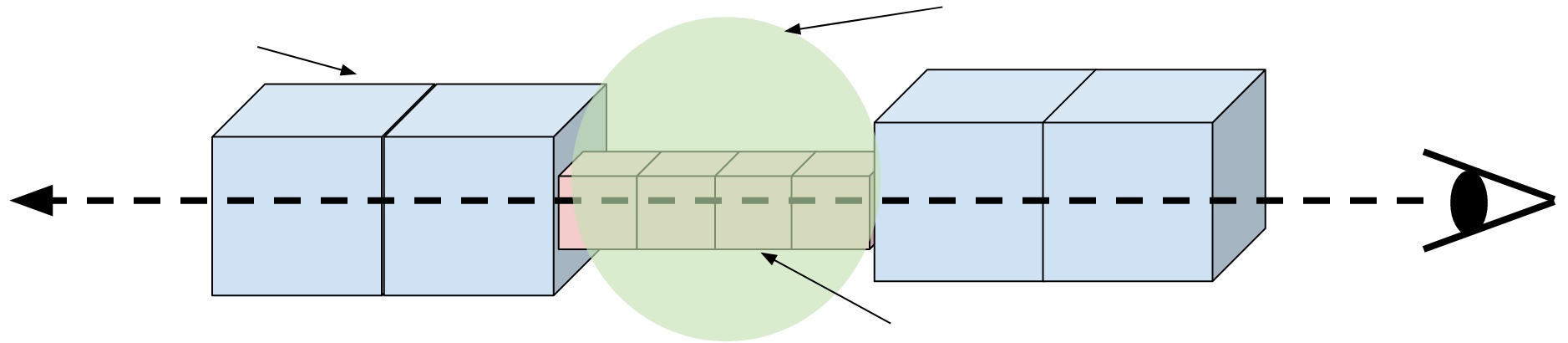}
    \put(-232, 45){\footnotesize Coarse Octree}
    \put(-92, 47){\footnotesize Attention Region}
    \put(-98, 1){\footnotesize Fine Octree}
    \vspace{-5pt}
    \caption{Dual-map ray casting implementation. When ray casting, the coarse and fine maps are used outside and inside the Attention Region respectively.}
    \label{fig:nbv_raycast}
    \vspace{-10pt}
\end{figure}

When ray casting, for both updating the environment model and calculating information gain, rays are cast through nodes in the coarse octree outside the Attention Region, and the fine octree inside the Attention Region, as demonstrated in Fig.~\ref{fig:nbv_raycast}. This significantly improves the computational cost of ray casting by reducing the number of traversed voxels when using higher map resolutions. 

While this approach could be implemented using a single octree, maintaining two maps provides more fine-grained control over the planning resolutions, while adding trivial additional overhead. Our implementation also easily scales to include multiple fine octrees for future work on more complete global coverage, as discussed in Section~\ref{sec:conclusion}.

\subsection{Viewpoint Sampling}
Our viewpoint sampling method is inspired by ROI targeted sampling presented in \cite{nbv_zaenker}. First, ROI frontier voxels within the sampling tree are identified and used as target candidates. A frontier is the region between an empty voxel and unknown space, and an ROI frontier is a frontier in the vicinity of a known ROI. To determine ROI frontiers, the 6-neighborhoods of all ROIs are checked for free nodes. For all resulting free nodes, their 6-neighborhoods are checked for unknown neighbors. All free nodes that have at least one unknown neighbor are used as viewpoint targets.
For each viewpoint target, candidate viewpoints are sampled from a partial Fibonacci sphere centered around the target at a specified sensor distance. Viewpoints that lie outside the workspace tree are discarded.

\subsection{Viewpoint Evaluation}\label{subsec:viewpoint_evaluation}
The information gain (IG) is estimated for all candidate viewpoints. For each viewpoint, multiple rays spanning the field of view of the camera are cast from the viewpoint along the direction towards the target using the method from Section~\ref{subsec:dual_map}. Rays terminate when they encounter an occupied voxel or exceed a maximum distance.  

Our IG metric is an attention-guided version of the Unobserved Voxel IG presented by \cite{nbv_zaenker}. The IG for ray $r \in R$ is the number of unknown voxels along the ray that lie in the Attention Region $N_{A, u, r}$ divided by the total number of nodes along the ray that are inside the Attention Region $N_{A, r}$. The IG is averaged across all rays (Eq.~\ref{eq:ig_0} and~\ref{eq:ig_1}). This attention-based approach ensures that only unknown voxels that lie within our sizing area of interest contribute to the information gain, which results in more informative planning.
\begin{equation}
    IG_r = 
    \begin{cases}
        \frac{N_{A, u, r}}{N_{A, r}},& \text{if } N_{A, r} > 0\\[5pt]
        0,              & \text{otherwise}
    \end{cases}
    \label{eq:ig_0}
\end{equation}
% \begin{equation}
%     IG_r = \frac{N_{A, u, r}}{N_{A, r}}
%     \label{eq:ig_0}
%     \vspace{-4pt}
% \end{equation}

\begin{equation}
    IG = \frac{1}{|R|}\sum_{r \in R} IG_r
    \label{eq:ig_1}
\end{equation}

We also compute a cost $C$ to move to the viewpoint. This is the Euclidean distance between the position of the viewpoint and the current camera position. The Euclidean distance is used because computing the joint trajectory for every viewpoint is time-consuming. The cost is scaled by a constant $\alpha$, and the final utility of the viewpoint is
\begin{equation} \label{utility_eq}
U = IG - \alpha \cdot C
\end{equation}

A max-heap is created from the viewpoints with order determined by their utility. The planner iterates through the heap until a viewpoint is found with a utility greater than a pre-determined threshold and that the motion planner can find a successful path to. If no viewpoints are left in the heap, new viewpoints are sampled.

% \begin{algorithm}[t]
%     \textbf{Parameter}: $u_{t}$
%     \begin{algorithmic}[1]
%         \For{\textit{Planning Duration}}
%             \State $r_s$ = roiTargetSample();
%             \State $h_s$ = makeHeap($r_s$);
%             \While {$h_s \neq \emptyset$ \textbf{and} peek($h_s$) $>$ $u_t$}
%                 \State vp $=$ pop($h_s$);
%                 \If {moveToPose(vp)}
%                     \State break;
%                 \EndIf
%             \EndWhile
%         \EndFor
%     \end{algorithmic}
%     \caption{Viewpoint Planning}
%     \label{alg:viewpoint_planning}
% \end{algorithm}

% \subsection{Viewpoint Selection}
% Our viewpoint planning algorithm is described in Algorithm~\ref{alg:viewpoint_planning}. Viewpoint candidates are sampled using ROI targeted sampling. A max-heap is created from the viewpoints with order determined by their utility. The planner iterates through the heap until a viewpoint is found with a utility greater than a pre-determined threshold and that the motion planner can find a successful path to. If no viewpoints are left in the heap, new viewpoints are sampled. The process of viewpoint planning and capturing images repeats until the desired planning duration is exceeded.

%%%%%%%%%%%%%%%%%%%%%%%%%%%%%%%%%%%%%%%%%%%%%%%%%%%%%%%%%%%%%%%%%%%%%%%%%%%%%%%%
\section{Sizing}\label{sec:nbv_sizing}

Prior methods size fruit using either occupancy information from the volumetric map~\cite{nbv_zaenker} or automated shape completion methods~\cite{dissim_zaenker, superellipse}. Neither of these approaches adapt well to smaller fruit. The former has limited precision, and the second is unreliable as it is difficult to capture enough of the fruit's surface. Additionally, both use simple Euclidean clustering methods that do not extend to fruits that are close to each another. Instead, we look to 2D photogrammetric approaches, which have been used to size small fruit in agriculture \cite{Qadri, gongal, olive, on_tree}. These methods directly estimate the widths and heights of fruit by fitting simple 2D geometric shapes. The challenge is associating fruitlet detections across each image so that the least occluded image can be used to size each fruit. We tackle the data association problem by first globally registering the point clouds using a robust estimation method, and solving for associations using Highly Connected Subgraph Clustering~\cite{hcs}.

\subsection{Global Registration}
\begin{figure}[ht]
    \centering
    \includegraphics[width=0.9\linewidth]{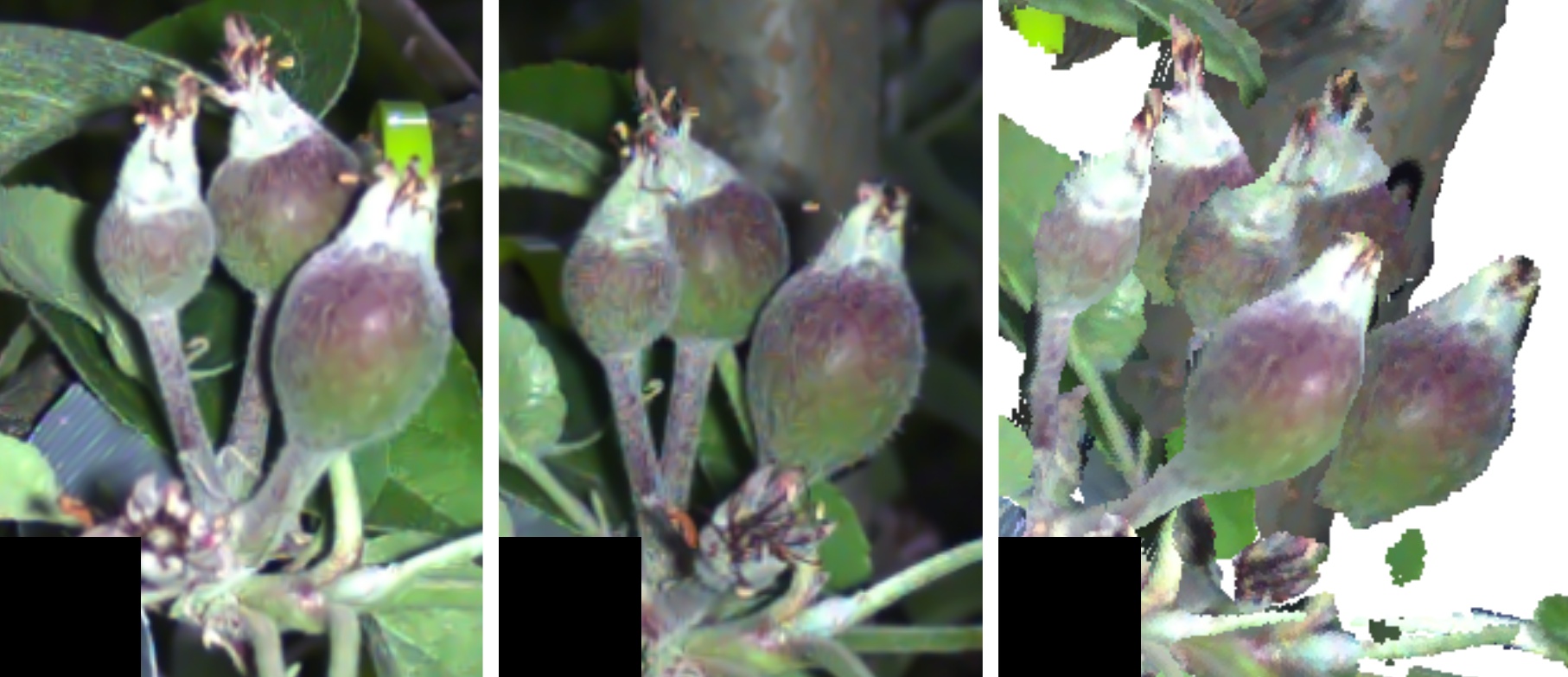}
    \put(-217, 7){\textcolor{white}{(a)}}
    \put(-146, 7){\textcolor{white}{(b)}}
    \put(-75, 7){\textcolor{white}{(c)}}
    \vspace{-6pt}
    \caption{Images (a) and (b) with similar camera poses have a noticeable offset in the point clouds (c) as a result of sensor noise and wind. }
    \label{fig:nbv_reproj_error}
    \vspace{-8pt}
\end{figure}

In each frame, there is an offset in the re-projected point cloud as a result of both sensor noise and wind, as visualized in Fig.~\ref{fig:nbv_reproj_error}. Qualitatively, we have observed translational shifts in the point clouds typically within the range of 3mm-5mm, but up to as large as 1cm. While this error may be negligible for larger fruit, is is non-trivial when sizing apple fruitlets, which are often only a few millimeters apart.

Using local methods for point cloud registration, such as Iterative Closest Point, will often fail as a result of the sparse viewpoints and inconsistent structure in the point clouds. Similarly, directly solving for partial assignment using methods such as the Hungarian Algorithm~\cite{Kuhn1955Hungarian} will not work because occlusions prevent every fruitlet from being detected in every image.

 To globally register the point clouds, we adopt a robust estimation approach. Robust estimation is a popular method used in Simultaneous Localization and Mapping to reduce the effect of outliers in the assumed model. In our case, outliers represent inconsistent detections across images as a result of occlusions and false positive detections. We use robust estimation to globally register all point clouds simultaneously. This is advantageous as fruitlet detections across the span of all images are more globally consistent than detections between individual frames.

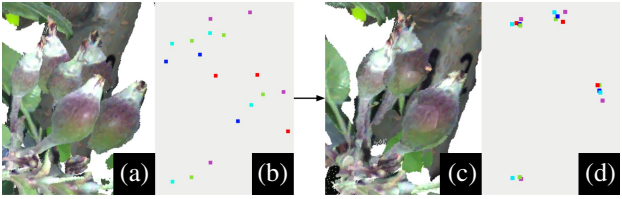
\begin{figure}[th]
    \centering
    \includegraphics[width=0.95\linewidth]{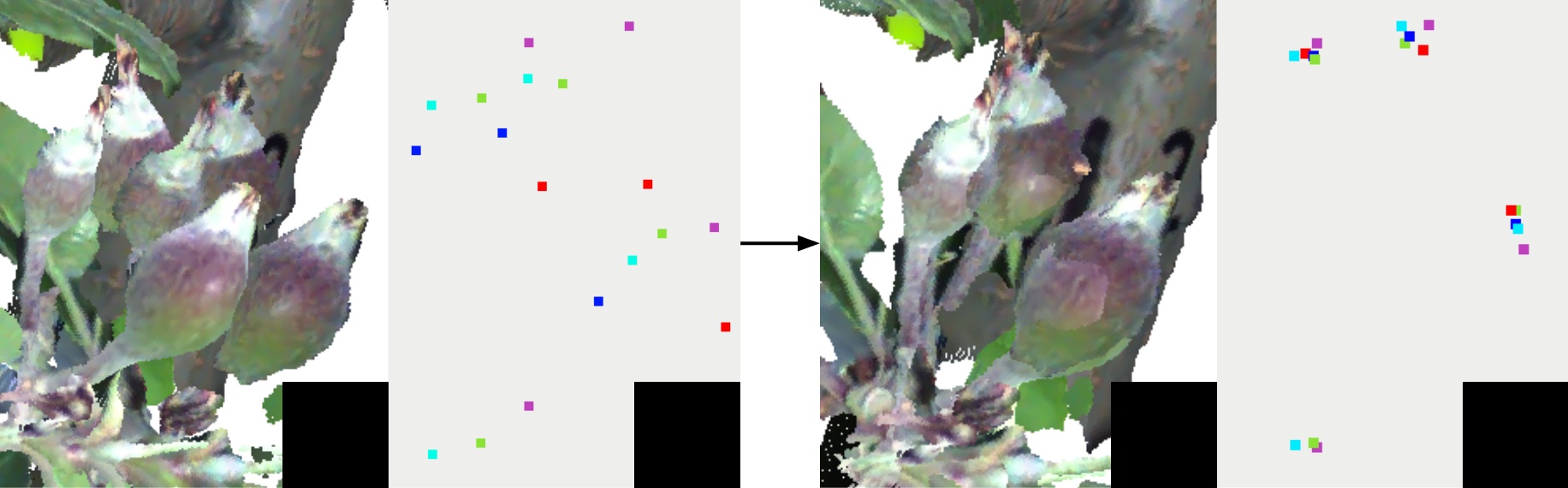}
    \put(-189, 5){\textcolor{white}{(a)}}
    \put(-137, 5){\textcolor{white}{(b)}}
    \put(-65, 5){\textcolor{white}{(c)}}
    \put(-13, 5){\textcolor{white}{(d)}}
    \vspace{-8pt}
    \caption{(a) and (c) Point clouds for two frames before and after global registration. (b) and (d) Fruitlet centroids before and after global registration for all frames where each color represents detections in a single image.}
    \vspace{-11pt}
    \label{fig:nbv_rob_est}
\end{figure}

 We formulate the task as a nonlinear least-squares problem. The objective is to find a transformation $\mathbf{T_i}$ for each frame $i \in \{1, ..., N\}$ that correctly aligns the point clouds. For each image $\mathbf{I}_i$, the centroid $c_{ij} \in \mathbb{R}^3$ of every fruitlet $s_{ij} \in \mathbf{S_i}$, $j \in \{1, ..., M_i\}$ is extracted, and we solve for the set of transformations $\mathbf{T} = \{\mathbf{T_1}, ..., \mathbf{T_N}\}$ that minimizes
\begin{equation}
    \underset{\mathbf{T}}{\text{minimize}} \frac{1}{2}\sum_{i=1}^N\sum_{j=i+1}^N\sum_{k=1}^{M_i}\sum_{l=1}^{M_j}\rho(|| \mathbf{T_i}c_{ik} - \mathbf{T_j}c_{jl}||^2)
\end{equation}
where $\rho$ is a scaled arctan function. The optimization problem is solved using a trust-region reflective algorithm~\cite{STIR, BYRD}. After solving for the refined transformations $\mathbf{T}$, the point clouds correctly align, as shown in Fig.~\ref{fig:nbv_rob_est}.

\subsection{HCS Clustering}
After globally registering the point clouds, we can associate the fruitlet detections across images. We need the association method to be robust to spurious and missed detections. To accomplish this, we utilize Highly Connected Subgraph (HCS) clustering~\cite{hcs}. HCS clustering is a recursive graph-clustering method that partitions a graph into subgraphs by taking the minimum cut. If the number of edges in the minimum cut is greater than half the number of nodes in the graph, the graph is determined to be highly connected and is a cluster. Otherwise, the edges in the minimum cut are removed, and the process repeats for each new subgraph.

\begin{figure}[ht]
    \centering
    \vspace{-5pt}
    \includegraphics[width=0.88\linewidth]{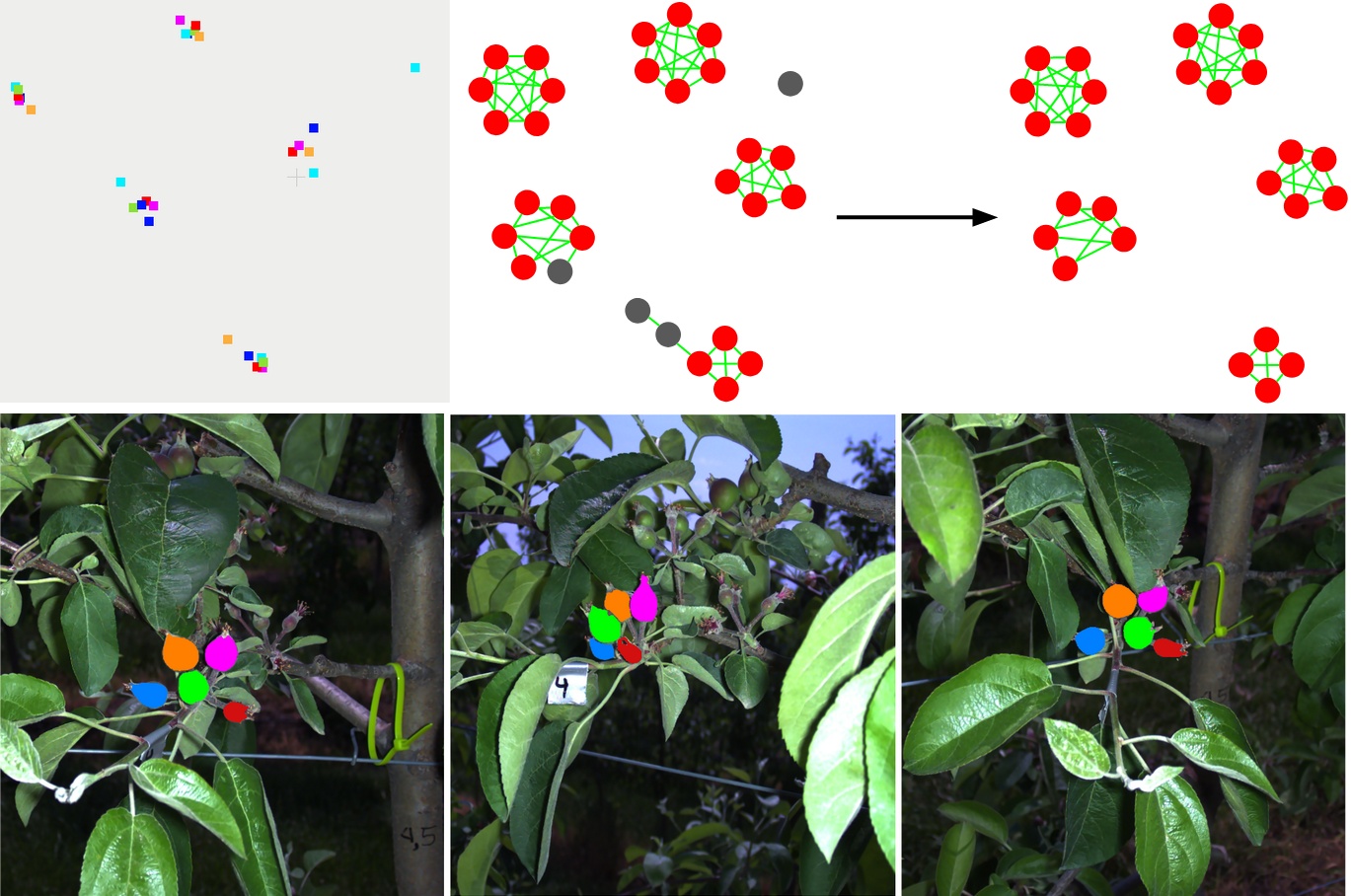}
    \put(-78, 101){\scriptsize HCS}
    \put(-85, 93){\scriptsize Clustering}
    \vspace{-5pt}
    \caption{Top: A point cloud of centroids is used to build a graph which consists of false detections (gray). HCS clustering removes the false detections and each subgraph represents an individual fruitlet. Bottom: Each color represents the same fruitlet associated in different images.}
    \vspace{-10pt}
    \label{fig:hcs_comb}
\end{figure}

We construct a graph $G(V, E)$ consisting of nodes $V$ and edges $E$. We treat the corrected world-transformed centroid of each detected fruitlet in each image as a node.
\begin{equation}
    V = \{\mathbf{T_i}c_{ij} \forall \ i \in \{1, ..., N\} \text{ and } j \in \{1, ..., M_i\}\}
\end{equation}
An edge $e_{ij}$ is created between nodes $v_i$ and $v_j$ if i) the detections corresponding to the centroids are from different images, and ii) the Euclidean distances between the nodes is less than a threshold $\tau$.
\begin{equation}
    E = \{e_{ij} \forall \ i, j \  \text{s.t. Im}(i) \neq \text{Im}(j) \wedge \lVert v_i - v_j \rVert < \tau\}
\end{equation}
Im$(i)$ represents the image corresponding to node $v_i$. The first edge requirement enforces that two detections in the same image cannot associate to the same fruitlet. The second edge requirement limits only reasonable associations. The graph is clustered using HCS clustering, and each resulting cluster represents a fruitlet association. A visualization of the process can be seen in Fig.~\ref{fig:hcs_comb}.

% %%%
% \newcolumntype{A}{>{\centering\arraybackslash\hsize=.1\hsize}X}
% \newcolumntype{B}{>{\centering\arraybackslash\hsize=.1\hsize}X}
% \newcolumntype{D}{>{\centering\arraybackslash\hsize=.1\hsize}X}
% \newcolumntype{E}{>{\centering\arraybackslash\hsize=.1\hsize}X}
% \newcolumntype{F}{>{\centering\arraybackslash\hsize=.1\hsize}X}
% \newcolumntype{G}{>{\centering\arraybackslash\hsize=.1\hsize}X}
% \newcolumntype{H}{>{\centering\arraybackslash\hsize=.1\hsize}X}
% \newcolumntype{I}{>{\centering\arraybackslash\hsize=.1\hsize}X}
% \begin{table*}
% \caption{Experimental Results}
% \vspace{-8pt}
% \centering
% \begin{tabularx}{\linewidth}{@{}ABDEFGHI@{}} 
% \hline\hline %inserts double horizontal lines
% \multirow{2}{*}{{\shortstack[c]{\vspace{5pt}\\Metric}}} &
% \multicolumn{6}{p{13cm}}{\centering Simulated} & \multicolumn{1}{p{1.8cm}}{\centering Real-World}\\
% \cmidrule(lr){2-7} \cmidrule(lr){8-8}
%  & FVP & SM-FVP &  NAG-FVP & RVP & RS &  EDS & FVP\\[0.5ex] % inserts table
% \hline
% MP (\%) & \textbf{95.8} & 90.7 & 87.5 & 95.4 & 88.4 & 85.8 & 88.1 \\
% MAE (mm) & \textbf{0.613} & 0.834 & 0.983 & 0.841 & 1.40 & 1.44 & 0.811 \\
% MAPE (\%) & \textbf{6.17} & 8.33 & 9.92 & 8.36 & 13.2 & 13.1 & 7.24 \\
% MNI & 7.07 & \textbf{TBD} & TBD & \textbf{TBD} & 7.65 & 6 & TBD \\
% $R^2$ & \textbf{0.907} & 0.824 & 0.746 & 0.767 & 0.519 & 0.524 & 0.909 \\
% \hline\hline
% \end{tabularx}
% \label{table:results}
% \vspace{-12pt}
% \end{table*}
% %%%
%%%
\newcolumntype{A}{>{\centering\arraybackslash\hsize=.1\hsize}X}
\newcolumntype{B}{>{\centering\arraybackslash\hsize=.1\hsize}X}
\newcolumntype{D}{>{\centering\arraybackslash\hsize=.1\hsize}X}
\newcolumntype{E}{>{\centering\arraybackslash\hsize=.1\hsize}X}
\newcolumntype{F}{>{\centering\arraybackslash\hsize=.1\hsize}X}
\newcolumntype{G}{>{\centering\arraybackslash\hsize=.1\hsize}X}
\newcolumntype{H}{>{\centering\arraybackslash\hsize=.1\hsize}X}
\newcolumntype{I}{>{\centering\arraybackslash\hsize=.1\hsize}X}
\newcolumntype{J}{>{\centering\arraybackslash\hsize=.1\hsize}X}
\newcolumntype{K}{>{\centering\arraybackslash\hsize=.1\hsize}X}
\begin{table*}
\vspace{2pt}
\caption{Experimental Results}
\vspace{-8pt}
\centering
\begin{tabularx}{\linewidth}{@{}ABDEFGHIJK@{}} 
\hline\hline %inserts double horizontal lines
\multirow{2}{*}{{\shortstack[c]{\vspace{5pt}\\Metric}}} &
\multicolumn{8}{p{13cm}}{\centering Simulated} & \multicolumn{1}{p{1.5cm}}{\centering Real-World}\\
\cmidrule(lr){2-9} \cmidrule(lr){10-10}
 & FVP (ours) &  NAG-FVP & SM-FVP-LR & SM-FVP-HR & RVP-LR & RVP-HR & RS &  EDS & FVP (ours)\\[0.5ex] % inserts table
\hline
MP (\%) & \textbf{95.8} & 87.5 & 90.7 & 88.0 & 95.4 & 80.0 & 88.4 & 85.8 & 88.1 \\
MAE (mm) & \textbf{0.613} & 0.983 & 0.834 & 1.43 & 0.841 & 1.44 & 1.40 & 1.44 & 0.811 \\
MAPE (\%) & \textbf{6.17} & 9.92 & 8.33 & 12.9 & 8.36 & 13.1 & 13.2 & 13.1 & 7.24 \\
$R^2$ & \textbf{0.907} & 0.746 & 0.824 & 0.534 & 0.767 & 0.575 & 0.519 & 0.524 & 0.909 \\
\hline\hline
\end{tabularx}
\label{table:results}
\vspace{-11pt}
\end{table*}
%%%

\subsection{Ellipse Fitting and Sizing}
\begin{figure}[t]
    \centering
    \includegraphics[width=0.88\linewidth]{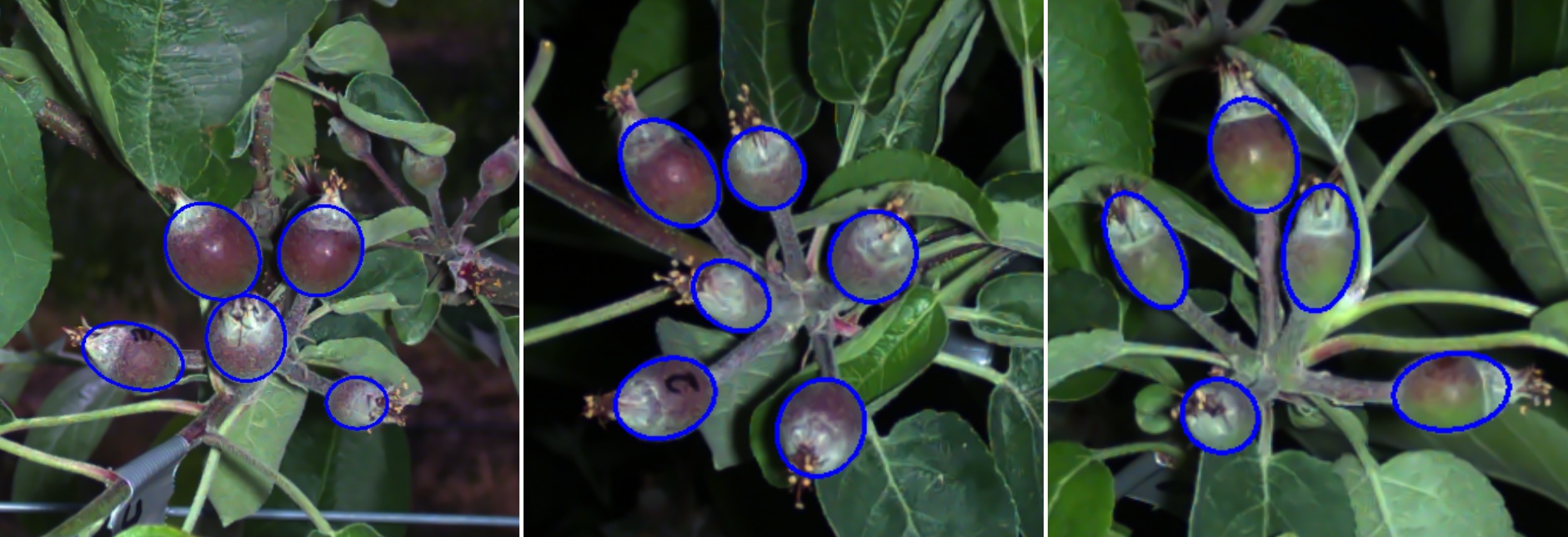}
    \vspace{-8pt}
    \caption{Examples of ellipses fit to fruitlets.}
    \vspace{-11pt}
    \label{fig:fruitlet_ellipse}
\end{figure}

For every fruitlet, an ellipse is fit using the OpenCV~\cite{opencv_library} fitEllipse function (Fig.~\ref{fig:fruitlet_ellipse}). The size is calculated as $\frac{b \times a_{\text{minor}}}{d_{\text{med}}}$, where $b$ is the baseline of the stereo camera, $d_{\text{med}}$ is the median disparity value of the segmented pixels, and $a_{\text{minor}}$ is the length of the minor axis of the fit ellipse. The derivation for this equation can be found in \cite{Qadri}. For each fruitlet, the final size is determined to be largest among its associations.

%%%%%%%%%%%%%%%%%%%%%%%%%%%%%%%%%%%%%%%%%%%%%%%%%%%%%%%%%%%%%%%%%%%%%%%%%%%%%%%%
\section{Experiments}

\subsection{Simulated Experiments}
\begin{figure}[th]
    \centering
    \includegraphics[width=0.9\linewidth]{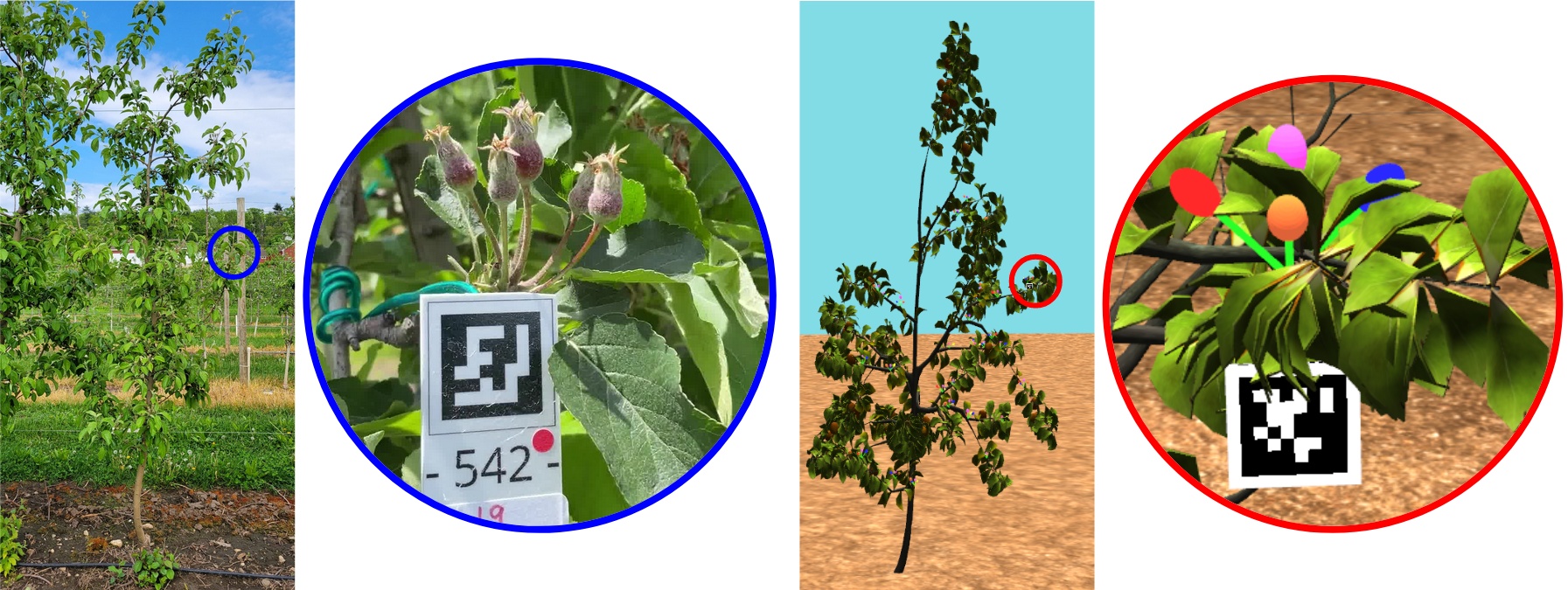}
    \vspace{-8pt}
    \caption{Example real and simulated apple trees and clusters.}
    \label{fig:nbv_sim_env}
    \vspace{-8pt}
\end{figure}

Experiments in simulation were performed on synthetic apple trees in a Gazebo~\cite{gazebo} simulated environment. Fruitlets are modelled as ellipsoids, and each fruitlet in a cluster is assigned a unique color. Three different structural topologies with varying levels of foliage were created using SpeedTree\footnote{\url{www.speedtree.com}}. For each tree, 16 trials were run with randomly generated clusters for a total of 48 trials. Clusters were randomly assigned 3-6 fruitlets with varying poses. The sizes of the fruitlets were also random, with diameters ranging from 7mm-14mm. For each trial, a single cluster is selected and AprilTag hung in close proximity to it, as shown in Fig.~\ref{fig:nbv_sim_env}.

The different coloring of the fruitlets is used only for instance segmentation and has no effect on data association. Fruitlets are segmented by HSV thresholding followed by Density-Based Spatial Clustering~\cite{ester1996densitybased} to separate different instances of the same color belonging to different clusters.

We assess the effectiveness of our approach by running ablations, comparing to two naive approaches, and comparing the planner presented by Zaenker \textit{et al.}~\cite{nbv_zaenker}. All planners were given a combined planning and execution time of two minutes to move around the world and capture images. For each experiment, coarse and fine octree resolutions of 1cm and 3mm were used. The methods evaluated include 

% \begin{enumerate}
%     \item 
%     \textit{Fruitlet Viewpoint Planner (FVP)}: Our method as described in Section~\ref{sec:viewpoint_planning}. Coarse and fine octree resolutions of 1cm and \todo{is it 3mm?}3mm were used.
%     \item
%     \textit{Single Map Fruitlet Viewpoint Planner (SM-FVP)}: The dual-map representation of the environment from Section \ref{subsec:dual_map} was removed and replaced with a single octree of 1cm resolution.
%     \item
%     \textit{Non-Attention-Guided Fruitlet Viewpoint Planner (NAG-FVP)}: The attention mechanism used to calculate information gain is removed. Coarse and fine octree resolutions of 1cm and \todo{is it 3mm?}3mm were used.  
%     \item 
%     \textit{ROI Viewpoint Planner (RVP)}: The ROI Viewpoint Planner presented by \cite{nbv_zaenker} with a 1cm octree resolution. 
%     %This method is equivalent to removing both the dual-map representation and the attention-guided utility as described by \textit{SM-FVP} and \textit{NAG-FVP}. 
%     \item
%     \textit{Random Sampler (RS)}: Viewpoints are randomly selected from a sphere surrounding the Attention Region.
%     \item 
%     \textit{Evenly Distributed Sampler (EDS)}:  \todo{is it 6?}6 viewpoints are sampled from a partial Fibonacci sphere surrounding the Attention Region.
% \end{enumerate}
\begin{enumerate}
    \item 
    \textit{Fruitlet Viewpoint Planner (FVP)}: Our method as described in Section~\ref{sec:viewpoint_planning}.
    \item
    \textit{Single Map Fruitlet Viewpoint Planner}: The dual-map representation of the environment from Section \ref{subsec:dual_map} was removed and replaced with a single coarse (\textit{SM-FVP-LR}) or fine (\textit{SM-FVP-HR}) octree.
    \item
    \textit{Non-Attention-Guided Fruitlet Viewpoint Planner (NAG-FVP)}: The attention mechanism used to calculate information gain is removed.  
    \item 
    \textit{ROI Viewpoint Planner}: The planner presented by \cite{nbv_zaenker} with one coarse (\textit{RVP-LR}) or fine (\textit{RVP-HR}) octree. 
    %This method is equivalent to removing both the dual-map representation and the attention-guided utility as described by \textit{SM-FVP} and \textit{NAG-FVP}. 
    \item
    \textit{Random Sampler (RS)}: Naive approach where viewpoints are randomly selected from a partial sphere surrounding the Attention Region.
    \item 
    \textit{Evenly Distributed Sampler (EDS)}: Naive approach where viewpoints are evenly sampled on a partial sphere surrounding the Attention Region.
\end{enumerate}

We report the match percent (MP), which is the percent of fruitlets that can be matched with ground truth. We also compare the mean absolute error (MAE) and mean absolute percentage error (MAPE) between the measured and ground truth sizes. The results can be seen in Table~\ref{table:results}.
\begin{equation}
    \text{MAE} = \frac{1}{N}\sum_{n=1}^{N} \left|\text{measured} - \text{gt}\right|
    \vspace{-3pt}
\end{equation}
\begin{equation}
    \text{MAPE} = \frac{100}{N}\sum_{n=1}^{N} \frac{\left|\text{measured} - \text{gt}\right|}{\text{gt}}
\end{equation}

\begin{figure}[!htbp]
    \centering
    \includegraphics[width=0.99\linewidth]{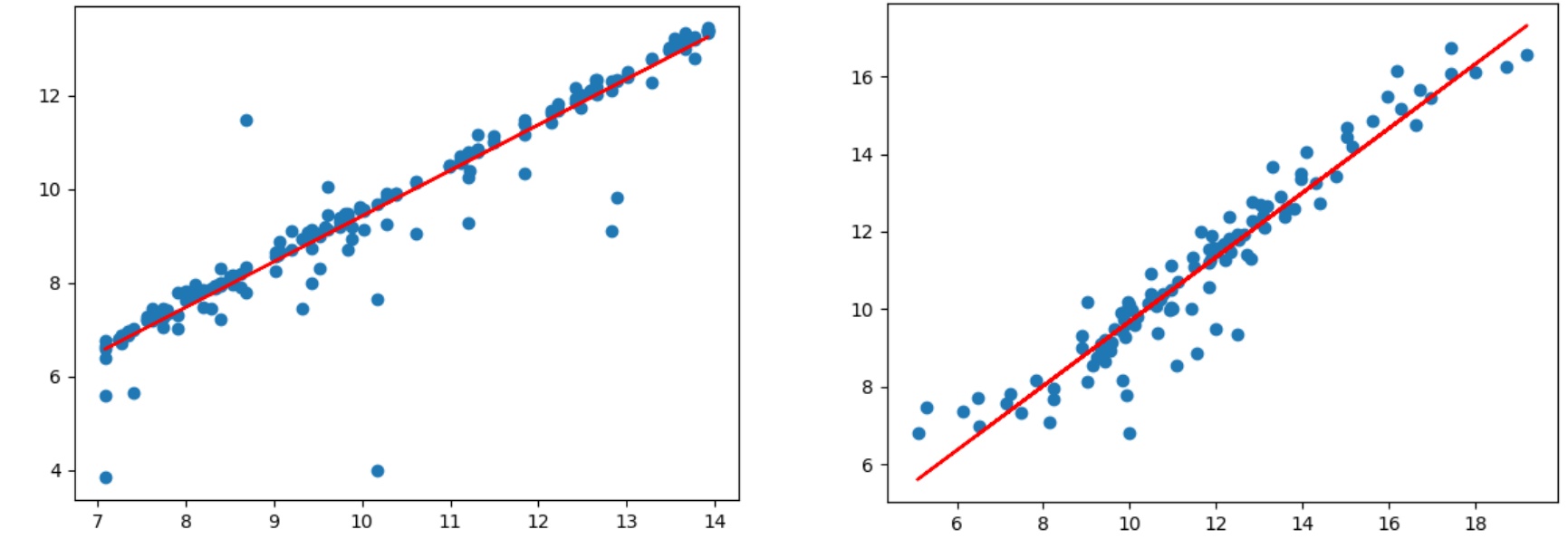}
    \put(-219, 86){\scriptsize FVP vs. Simulated Sizes}
    \put(-92, 86){\scriptsize FVP vs. Real-World Sizes}
    \put(-245, 12){\scriptsize \rotatebox{90}{Predicted Sizes (mm)}}
    \put(-120, 12){\scriptsize \rotatebox{90}{Predicted Sizes (mm)}}
    \put(-216, -6){\scriptsize {Ground Truth Sizes (mm)}}
    \put(-92, -6){\scriptsize {Ground Truth Sizes (mm)}}
    \put(-227, 73){\scriptsize {$R^2 = 0.907$}}
    \put(-100, 73){\scriptsize {$R^2 = 0.909$}}
    %\vspace{-1pt}
    \caption{Linear fit between predicted and ground truth sizes for our FVP in both simulation (left) and real-world (right) experiments.}
    \label{fig:nbv_r2}
    \vspace{-13pt}
\end{figure}
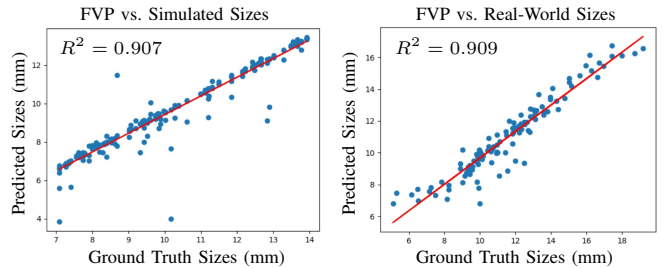

Our FVP achieves the highest MP, the lowest MAE, and the lowest MAPE compared to all other evaluated methods. Our planner matches 95.8\% of fruitlets and achieves a MAPE of 6.17\%. Additionally, we obtain the best linear fit with ground truth, with an $R^2$ score of 0.907 (Fig.~\ref{fig:nbv_r2}).

Analyzing the ablations and state of the art, SM-FVP-HR and RVP-HR performed the worst with regards to MAE, MAPE, and $R^2$ scores. The results were comparable to the naive approaches. This is because of the computational costs associated with using higher resolution octrees which result in a fewer number of evaluated viewpoints. NAG-FVP also did not perform well, again because of the additional costs required to update the fine octree without the benefits of attention-guided information gain. 

There is a trade-off between RVP-LR and SM-FVP-LR. RVP-LR is able to size more fruitlets compared to SM-FVP-LR, but has slightly larger MAE and MAPE. This could be explained by RVP-LR being encouraged to explore more, as voxels outside the Attention Region contribute the information gain. Because SM-FVP-LR applies the attention mechanism, it is incentivized to more accurately explore previously detected ROIs inside the Attention Region. 

Based on these results, our FVP effectively demonstrates that combining our dual-map representation with attention improves sizing performance across the presented metrics.

\subsection{Real-World Experiments}
Real-world experiments were conducted at the University of Massachusetts Amherst Cold Spring Orchard. The robotic system from Section \ref{sec:sys_overview} was used to size 30 McIntosh apple clusters on 05/22/2023 using our Fruitlet Viewpoint Planner. Caliper measurements were also recorded to be used as ground truth. The Mask-RCNN network was trained using 600 images of Fuji, Gala, and Honeycrisp clusters.

% \begin{table}[!htbp]
% \centering
% \caption{Experimental results on real data for our FVP.}
% \vspace{-5pt}
% \begin{tabular}{|c|c|}
% \hline
% \textbf{MP} (\%) & 88.1 \\
% \hline
% \textbf{MAE} (mm) & 0.811 \\
% \hline
% \textbf{MAPE} (\%) & 7.24 \\
% \hline
% \end{tabular}
% \label{tab:nbv_error_met_2}
% \vspace{-6pt}
% \end{table}

The results can be seen in Table~\ref{table:results}. Our Fruitlet Viewpoint Planner demonstrates impressive performance on the real-world experiments. The planner achieves a mean average percentage error of 7.24\% and an $R^2$ score of 0.909 between the predicted and caliper-measured sizes (Fig.~\ref{fig:nbv_r2}). 
In addition, 88.1\% percent of fruitlets were able to be matched. While this is a strong result, there is room for improvement. Through qualitative inspection, many missed instances were a result of the Mask R-CNN network missing detections. This is likely because no McIntosh apples were used in training. With additional training data, the performance of our planner with regards to this metric would likely improve.

%%%%%%%%%%%%%%%%%%%%%%%%%%%%%%%%%%%%%%%%%%%%%%%%%%%%%%%%%%%%%%%%%%%%%%%%%%%%%%%%
\section{Conclusion}\label{sec:conclusion}
Our next-best-view planner outperforms previous state of the art in simulation and demonstrates strong results on a real-world dataset. Although initially designed to size apple fruitlets, the presented methods could be used to size other small fruits and plants of comparable size. For future work, we hope to integrate more global viewpoint coverage to size multiple clusters at a time. Instead of using an AprilTag to identify a cluster, global broad scans of trees could be used to find high-density fruit regions using the methods presented in Section~\ref{tree_gen}. Our map representation could be extended to include multiple fine octrees and Attention Regions to capture finer coverage of each cluster.

%%%%%%%%%%%%%%%%%%%%%%%%%%%%%%%%%%%%%%%%%%%%%%%%%%%%%%%%%%%%%%%%%%%%%%%%%%%%%%%%
\section{Acknowledgements}
This work was partially supported by NSF/USDA NIFA 2020-01469-1022394, NSF Robust Intelligence 1956163, and NSF/USDA NIFA AIIRA AI Research Institute 2021-67021-35329. The authors would like to thank Abhi Silwal and Zack Rubinstein for their support throughout this project.

%This work was partially supported by NSF/USDA NIFA 2020-01469-1022394, NSF Robust Intelligence 1956163, and NSF/USDA NIFA AIIRA AI Research Institute 2021-67021-35329. The authors would like to thank Abhi Silwal, Zack Rubinstein, Dan Cooley, Jon Clements, and the University of Massachusetts Amherst Cold Spring Orchard for their assistance and support during field testing.

\clearpage
\newpage

\bibliographystyle{IEEEtran} % use IEEEtran.bst style
\bibliography{mybib}

% Generated by IEEEtran.bst, version: 1.14 (2015/08/26)
\begin{thebibliography}{10}
\providecommand{\url}[1]{#1}
\csname url@samestyle\endcsname
\providecommand{\newblock}{\relax}
\providecommand{\bibinfo}[2]{#2}
\providecommand{\BIBentrySTDinterwordspacing}{\spaceskip=0pt\relax}
\providecommand{\BIBentryALTinterwordstretchfactor}{4}
\providecommand{\BIBentryALTinterwordspacing}{\spaceskip=\fontdimen2\font plus
\BIBentryALTinterwordstretchfactor\fontdimen3\font minus \fontdimen4\font\relax}
\providecommand{\BIBforeignlanguage}[2]{{%
\expandafter\ifx\csname l@#1\endcsname\relax
\typeout{** WARNING: IEEEtran.bst: No hyphenation pattern has been}%
\typeout{** loaded for the language `#1'. Using the pattern for}%
\typeout{** the default language instead.}%
\else
\language=\csname l@#1\endcsname
\fi
#2}}
\providecommand{\BIBdecl}{\relax}
\BIBdecl

\bibitem{greene2013development}
D.~W. Greene, A.~N. Lakso, T.~L. Robinson, and P.~Schwallier, ``Development of a fruitlet growth model to predict thinner response on apples,'' \emph{HortScience}, vol.~48, no.~5, pp. 584--587, 2013.

\bibitem{fixed_apple}
\BIBentryALTinterwordspacing
T.~Hondo, K.~Kobayashi, and Y.~Aoyagi, ``Real-time prediction of growth characteristics for individual fruits using deep learning,'' \emph{Sensors}, vol.~22, no.~17, 2022. [Online]. Available: \url{https://www.mdpi.com/1424-8220/22/17/6473}
\BIBentrySTDinterwordspacing

\bibitem{taim}
S.~Kelly, A.~Riccardi, E.~Marks, F.~Magistri, T.~Guadagnino, M.~Chli, and C.~Stachniss, ``Target-aware implicit mapping for agricultural crop inspection,'' in \emph{2023 IEEE International Conference on Robotics and Automation (ICRA)}, 2023, pp. 9608--9614.

\bibitem{3dmovetosee}
C.~Lehnert, D.~Tsai, A.~Eriksson, and C.~McCool, ``3d move to see: Multi-perspective visual servoing towards the next best view within unstructured and occluded environments,'' in \emph{2019 IEEE/RSJ International Conference on Intelligent Robots and Systems (IROS)}, 2019, pp. 3890--3897.

\bibitem{counting_apples}
P.~Roy and V.~Isler, ``Active view planning for counting apples in orchards,'' in \emph{2017 IEEE/RSJ International Conference on Intelligent Robots and Systems (IROS)}, 2017, pp. 6027--6032.

\bibitem{deep_learn_nbv}
C.~Wu, R.~Zeng, J.~Pan, C.~C.~L. Wang, and Y.-J. Liu, ``Plant phenotyping by deep-learning-based planner for multi-robots,'' \emph{IEEE Robotics and Automation Letters}, vol.~4, no.~4, pp. 3113--3120, 2019.

\bibitem{PCN}
W.~Yuan, T.~Khot, D.~Held, C.~Mertz, and M.~Hebert, ``Pcn: Point completion network,'' in \emph{2018 International Conference on 3D Vision (3DV)}, 2018, pp. 728--737.

\bibitem{dec_mcts}
F.~Sukkar, G.~Best, C.~Yoo, and R.~Fitch, ``Multi-robot region-of-interest reconstruction with dec-mcts,'' in \emph{2019 International Conference on Robotics and Automation (ICRA)}, 2019, pp. 9101--9107.

\bibitem{decmcts}
\BIBentryALTinterwordspacing
G.~Best, O.~M. Cliff, T.~Patten, R.~R. Mettu, and R.~Fitch, ``Dec-mcts: Decentralized planning for multi-robot active perception,'' \emph{The International Journal of Robotics Research}, vol.~38, no. 2-3, pp. 316--337, 2019. [Online]. Available: \url{https://doi.org/10.1177/0278364918755924}
\BIBentrySTDinterwordspacing

\bibitem{nbv_zaenker}
\BIBentryALTinterwordspacing
T.~Zaenker, C.~Smitt, C.~McCool, and M.~Bennewitz, ``Viewpoint planning for fruit size and position estimation,'' \emph{CoRR}, vol. abs/2011.00275, 2020. [Online]. Available: \url{https://arxiv.org/abs/2011.00275}
\BIBentrySTDinterwordspacing

\bibitem{combglobloc}
T.~Zaenker, C.~Lehnert, C.~McCool, and M.~Bennewitz, ``Combining local and global viewpoint planning for fruit coverage,'' in \emph{2021 European Conference on Mobile Robots (ECMR)}, 2021, pp. 1--7.

\bibitem{dissim_zaenker}
R.~Menon, T.~Zaenker, N.~Dengler, and M.~Bennewitz, ``Nbv-sc: Next best view planning based on shape completion for fruit mapping and reconstruction,'' 2023.

\bibitem{burusa2022attentiondriven}
A.~K. Burusa, E.~J. van Henten, and G.~Kootstra, ``Attention-driven active vision for efficient reconstruction of plants and targeted plant parts,'' 2022.

\bibitem{apriltag}
E.~Olson, ``Apriltag: A robust and flexible visual fiducial system,'' in \emph{2011 IEEE International Conference on Robotics and Automation}, 2011, pp. 3400--3407.

\bibitem{flashcam2021}
A.~Silwal, T.~Parhar, F.~Yandun, H.~Baweja, and G.~Kantor, ``A robust illumination-invariant camera system for agricultural applications,'' in \emph{2021 IEEE/RSJ International Conference on Intelligent Robots and Systems (IROS)}, 2021, pp. 3292--3298.

\bibitem{silwal2021bumblebee}
\BIBentryALTinterwordspacing
A.~Silwal, F.~Yand{\'{u}}n, A.~K. Nellithimaru, T.~Bates, and G.~Kantor, ``Bumblebee: {A} path towards fully autonomous robotic vine pruning,'' \emph{CoRR}, vol. abs/2112.00291, 2021. [Online]. Available: \url{https://arxiv.org/abs/2112.00291}
\BIBentrySTDinterwordspacing

\bibitem{maskrcnn2017}
\BIBentryALTinterwordspacing
K.~He, G.~Gkioxari, P.~Doll{\'{a}}r, and R.~B. Girshick, ``Mask {R-CNN},'' \emph{CoRR}, vol. abs/1703.06870, 2017. [Online]. Available: \url{http://arxiv.org/abs/1703.06870}
\BIBentrySTDinterwordspacing

\bibitem{lipson2021raft}
L.~Lipson, Z.~Teed, and J.~Deng, ``{RAFT-Stereo: Multilevel Recurrent Field Transforms for Stereo Matching},'' \emph{arXiv preprint arXiv:2109.07547}, 2021.

\bibitem{moveit}
\BIBentryALTinterwordspacing
D.~Coleman, I.~A. Sucan, S.~Chitta, and N.~Correll, ``Reducing the barrier to entry of complex robotic software: a moveit! case study,'' \emph{CoRR}, vol. abs/1404.3785, 2014. [Online]. Available: \url{http://arxiv.org/abs/1404.3785}
\BIBentrySTDinterwordspacing

\bibitem{hcs}
E.~Hartuv and R.~Shamir, ``A clustering algorithm based on graph connectivity,'' \emph{Information Processing Letters}, vol.~76, pp. 175--181, 12 2000.

\bibitem{sorghumscan}
H.~Freeman, E.~Schneider, C.~H. Kim, M.~Lee, and G.~Kantor, ``3d reconstruction-based seed counting of sorghum panicles for agricultural inspection,'' in \emph{2023 IEEE International Conference on Robotics and Automation (ICRA)}, 2023, pp. 9594--9600.

\bibitem{octomap}
\BIBentryALTinterwordspacing
A.~Hornung, K.~M. Wurm, M.~Bennewitz, C.~Stachniss, and W.~Burgard, ``{OctoMap}: An efficient probabilistic {3D} mapping framework based on octrees,'' \emph{Autonomous Robots}, 2013, software available at \url{https://octomap.github.io}. [Online]. Available: \url{https://octomap.github.io}
\BIBentrySTDinterwordspacing

\bibitem{superellipse}
S.~Marangoz, T.~Zaenker, R.~Menon, and M.~Bennewitz, ``Fruit mapping with shape completion for autonomous crop monitoring,'' in \emph{2022 IEEE 18th International Conference on Automation Science and Engineering (CASE)}, 2022, pp. 471--476.

\bibitem{Qadri}
M.~Qadri, ``Robotic vision for 3d modeling and sizing in agriculture,'' Master's thesis, Carnegie Mellon University, Pittsburgh, PA, August 2021.

\bibitem{gongal}
\BIBentryALTinterwordspacing
A.~Gongal, M.~Karkee, and S.~Amatya, ``Apple fruit size estimation using a 3d machine vision system,'' \emph{Information Processing in Agriculture}, vol.~5, no.~4, pp. 498--503, 2018. [Online]. Available: \url{https://www.sciencedirect.com/science/article/pii/S2214317317302408}
\BIBentrySTDinterwordspacing

\bibitem{olive}
J.~M. Ponce, A.~Aquino, B.~Millan, and J.~M. Andújar, ``Automatic counting and individual size and mass estimation of olive-fruits through computer vision techniques,'' \emph{IEEE Access}, vol.~7, pp. 59\,451--59\,465, 2019.

\bibitem{on_tree}
\BIBentryALTinterwordspacing
Z.~Wang, K.~B. Walsh, and B.~Verma, ``On-tree mango fruit size estimation using rgb-d images,'' \emph{Sensors}, vol.~17, no.~12, 2017. [Online]. Available: \url{https://www.mdpi.com/1424-8220/17/12/2738}
\BIBentrySTDinterwordspacing

\bibitem{Kuhn1955Hungarian}
H.~W. Kuhn, ``{The Hungarian Method for the Assignment Problem},'' \emph{Naval Research Logistics Quarterly}, vol.~2, no. 1--2, pp. 83--97, March 1955.

\bibitem{STIR}
\BIBentryALTinterwordspacing
M.~A. Branch, T.~F. Coleman, and Y.~Li, ``A subspace, interior, and conjugate gradient method for large-scale bound-constrained minimization problems,'' \emph{SIAM Journal on Scientific Computing}, vol.~21, no.~1, pp. 1--23, 1999. [Online]. Available: \url{https://doi.org/10.1137/S1064827595289108}
\BIBentrySTDinterwordspacing

\bibitem{BYRD}
\BIBentryALTinterwordspacing
R.~H. Byrd, R.~B. Schnabel, and G.~A. Shultz, ``Approximate solution of the trust region problem by minimization over two-dimensional subspaces,'' \emph{Math. Program.}, vol.~40, no. 1-3, pp. 247--263, 1988. [Online]. Available: \url{https://doi.org/10.1007/BF01580735}
\BIBentrySTDinterwordspacing

\bibitem{opencv_library}
G.~Bradski, ``{The OpenCV Library},'' \emph{Dr. Dobb's Journal of Software Tools}, 2000.

\bibitem{gazebo}
N.~Koenig and A.~Howard, ``Design and use paradigms for gazebo, an open-source multi-robot simulator,'' in \emph{2004 IEEE/RSJ International Conference on Intelligent Robots and Systems (IROS) (IEEE Cat. No.04CH37566)}, vol.~3, 2004, pp. 2149--2154 vol.3.

\bibitem{ester1996densitybased}
M.~Ester, H.-P. Kriegel, J.~Sander, and X.~Xu, ``A density-based algorithm for discovering clusters in large spatial databases with noise,'' in \emph{Proc. of 2nd International Conference on Knowledge Discovery and}, 1996, pp. 226--231.

\end{thebibliography}

\end{document}